\documentclass{article}

\usepackage{microtype}
\usepackage{graphicx}
\usepackage{subfigure}
\usepackage{booktabs} 

\usepackage{hyperref}

\usepackage[accepted]{mlsys2022}

\usepackage[utf8]{inputenc} 
\usepackage[T1]{fontenc}    
\usepackage{hyperref}       
\usepackage{url}            
\usepackage{booktabs}       
\usepackage{amsfonts}       
\usepackage{nicefrac}       
\usepackage{microtype}      
\usepackage{xcolor}         
\usepackage{graphicx}
\usepackage{amsmath,amssymb,amsfonts}
\usepackage{caption}
\usepackage[export]{adjustbox}

\usepackage{array}
\newcommand{\PreserveBackslash}[1]{\let\temp=\\#1\let\\=\temp}
\newcolumntype{C}[1]{>{\PreserveBackslash\centering}p{#1}}
\newcolumntype{R}[1]{>{\PreserveBackslash\raggedleft}p{#1}}
\newcolumntype{L}[1]{>{\PreserveBackslash\raggedright}p{#1}}

\newcommand{\IGNORE}[1]{}

\newcommand\codecomment[1]{{\color{gray}#1}}

\mlsystitlerunning{A Transferable Approach for Partitioning Machine Learning Models on Multi-Chip-Modules}

\begin{document}

\twocolumn[
\mlsystitle{A Transferable Approach for Partitioning Machine Learning Models on Multi-Chip-Modules}



\mlsyssetsymbol{equal}{*}

\begin{mlsysauthorlist}
\mlsysauthor{Xinfeng Xie}{google,ucsb}
\mlsysauthor{Prakash Prabhu}{google}
\mlsysauthor{Ulysse Beaugnon}{google}
\mlsysauthor{Phitchaya Mangpo Phothilimthana}{google}
\mlsysauthor{Sudip Roy}{google}
\mlsysauthor{Azalia Mirhoseini}{google}
\mlsysauthor{Eugene Brevdo}{google}
\mlsysauthor{James Laudon}{google}
\mlsysauthor{Yanqi Zhou}{google}
\end{mlsysauthorlist}

\mlsysaffiliation{google}{Google}
\mlsysaffiliation{ucsb}{School of Electrical and Computer Engineering, University of California, Santa Barbara, California, United States}

\mlsyscorrespondingauthor{Xinfeng Xie}{xinfeng@ucsb.edu}
\mlsyscorrespondingauthor{Yanqi Zhou}{yanqiz@google.com}

\mlsyskeywords{ML Compiler, Model Partitioning, ML Accelerator, Multi-Chip-Modules, Reinforcement Learning}

\vskip 0.3in

\begin{abstract}
Multi-Chip-Modules (MCMs) reduce the design and fabrication cost of machine learning (ML) accelerators while delivering performance and energy efficiency on par with a monolithic large chip. 
However, ML compilers targeting  MCMs need to solve complex optimization problems optimally and efficiently to achieve this high performance. 
One such problem is the \emph{multi-chip partitioning} problem where compilers determine the optimal partitioning and placement of operations in tensor computation graphs on chiplets in MCMs. 
Partitioning ML graphs for MCMs is particularly hard as the search space grows exponentially with the number of chiplets available and the number of nodes in the neural network. 
Furthermore, the constraints imposed by the underlying hardware produce a search space where valid solutions are extremely sparse. 
In this paper, we present a strategy using a deep reinforcement learning (RL) framework to emit a possibly invalid candidate partition that is then corrected by a constraint solver. 
Using the constraint solver ensures that RL encounters valid solutions in the sparse space frequently enough to converge with fewer samples as compared to non-learned strategies.
The architectural choices we make for the policy network allow us to generalize across different ML graphs.
Our evaluation of a production-scale model, BERT, on real hardware reveals that the partitioning generated using RL policy achieves 6.11\% and 5.85\% higher throughput than random search and simulated annealing. In addition, fine-tuning the pre-trained RL policy reduces the search time from 3 hours to only 9 minutes, while achieving the same throughput as training RL policy from scratch.
\end{abstract}
]



\printAffiliationsAndNotice{}  

\section{Introduction}
Large machine learning models trained over big data have demonstrated promising results in tasks across many domains including computer vision ~\cite{mahajan2018exploring, he2015deep, ghiasi2019nasfpn}, speech~\cite{zhang2020pushing, gulati2020conformer, li2019neural}, and natural language processing~\cite{BERT, kaplan2020scaling, raffel2020exploring, amodei2015deep}. Training and inference for these models is an extremely compute-intensive process and is driving innovation across all layers of the stack to reduce the total cost of ownership of ML accelerators. 

Recent industrial prototypes of ML accelerators, such as Simba~\cite{shao2019simba} and multi-chip TPUs~\cite{dasari2021apparatus}, have demonstrated that multi-chip-modules (MCM) can reduce design and fabrication costs while delivering a similar performance and energy efficiency as that of a monolithic chip.
Instead of a large monolithic chip, MCM designs are composed of a set of small chips integrated into a package joined by off-chip interconnects~\cite{arunkumar2017mcm, kannan2015enabling, shao2019simba}.
MCM packages reduce the design cost as designing a smaller chip is easier than designing a monolithic large chip, and also have lower fabrication costs since the yield of chips is higher due to a smaller chip area~\cite{shao2019simba}. 

Since the memory sizes of individual chiplets in an MCM are significantly smaller than monolithic accelerators, training and inference of large ML models invariably require partitioning the dataflow graph of tensor computations over the chiplets.
Multi-chip partitioning is the problem of finding an assignment of operations in a computational graph to chiplets to maximize some performance metrics, typically throughput. 
The unique hardware characteristics of MCMs make the performance of ML models particularly sensitive to the quality of the partitioning, and therefore finding a good partitioning is an important optimization step in compilers targeting MCMs.

\begin{figure*}
    \centering
    \includegraphics[width=0.80\textwidth]{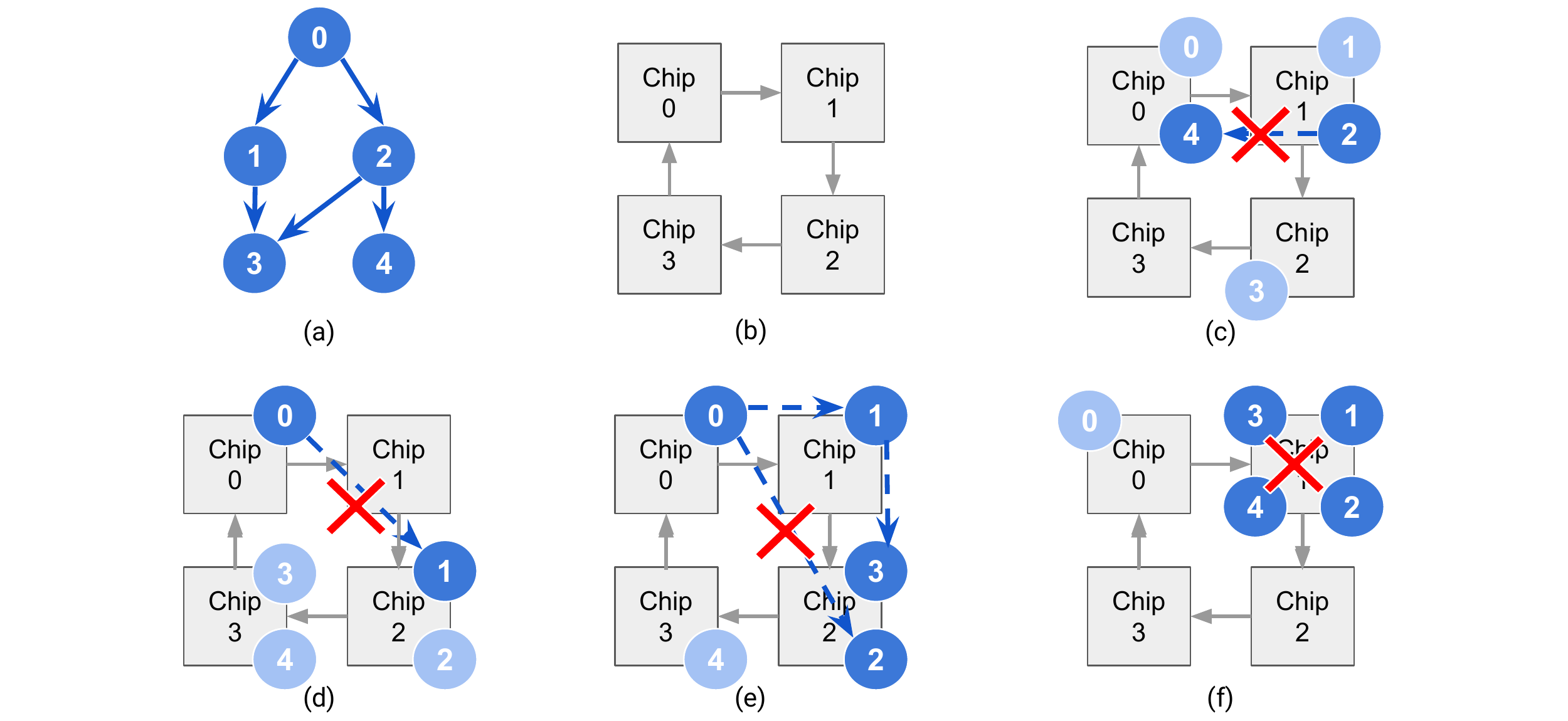}
    \caption{Examples of (a) a computation graph representing an ML workload, (b) multiple chips connected by uni-directional links, (c) an invalid partition violating the acyclic dataflow constraint, (d) an invalid partition violating the rule of no skipping chips, (e) an invalid partition violating the triangle dependency constraints, and (f) an invalid partition violating the memory allocation constraint.}
    \vspace{-1em}
    \label{fig:placement_problem}
    
\end{figure*}

Multi-chip partitioning is challenging for three reasons.
First, the specialized hardware architecture imposes constraints that invalidate large parts of the solution space. 
For example, in a multi-chip TPU package~\cite{dasari2021apparatus}, valid partitionings must assign operations to chiplets such that dataflows between chiplets are consistent with their relative positions along a uni-directional inter-ring network (Figure~\ref{fig:placement_problem}c).
Second, determining whether a partitioning is feasible or not requires executing the subsequent stages in the compilation process. 
For example, checking whether the peak memory usage for a particular placement is less than the available chiplet memory (Figure~\ref{fig:placement_problem}f) requires knowledge of the order of scheduling of operations that is only determined at a later compilation pass.
Finally, ML compilers usually have stringent time budgets for end-to-end compilation making it harder to find good partitionings in this sparse search space.

There are several existing solutions for the multi-chip partitioning task, such as constraint solvers, compiler heuristics, search-based algorithms, and reinforcement learning.
Unfortunately, these solutions can hardly find the optimal partition solution under strong constraints within a stringent time budget in the multi-chip setting.
It is a common approach to use constraint solvers in compilers to solve well-formulated optimization problems, such as loop transformations~\cite{bondhugula2008pluto}. 
After encoding the problem as a combinatorial decision optimization, off-the-shelf solvers are then applied to find the solution that minimizes an objective function \cite{feautrier1992some,bondhugula2008pluto,wilhelm2008worst,schoeberl2012statically}. 
Unfortunately, some dynamic constraints, such as memory allocation that happens later during the compilation process, are hard to be formulated at the stage of multi-chip partitioning. 
Moreover, objective functions with a closed-form formulation expressed in a solver logic typically fail to encapsulate the complexity of the MCM system.
On the other hand, hand-crafted compiler heuristics, such as greedy algorithms~\cite{shao2019simba} and dynamic programming~\cite{ding2021ios}, are frequently used in production compilers. 
Although they can search for valid partitions efficiently, they often fail to find the optimal placement due to their over-simplification of the performance model. 
Third, search-based compiler optimizations, such as random search and simulated annealing, can mitigate the problem of inaccurate performance models in real systems by sampling and evaluating partition candidates. 
However, these optimizations cannot find good candidates in a timely manner when the solution space is overwhelmingly large. 
In addition, search-based methods do not use prior knowledge to optimize an unseen new graph, thus they always search from scratch resulting in a long search time.
Recent work~\cite{zhou2021transferable} shows that Reinforcement Learning (RL) can be used to efficiently solve graph optimization problems that arise in ML compilers. 
However, the strict constraints of multi-chip partitioning invalidate most of the decisions in the action space. 
As a result, conventional RL methods fail due to insufficient valid samples or rewards. 

\begin{figure}[!t]
    \centering
    \includegraphics[width=1.0\linewidth]{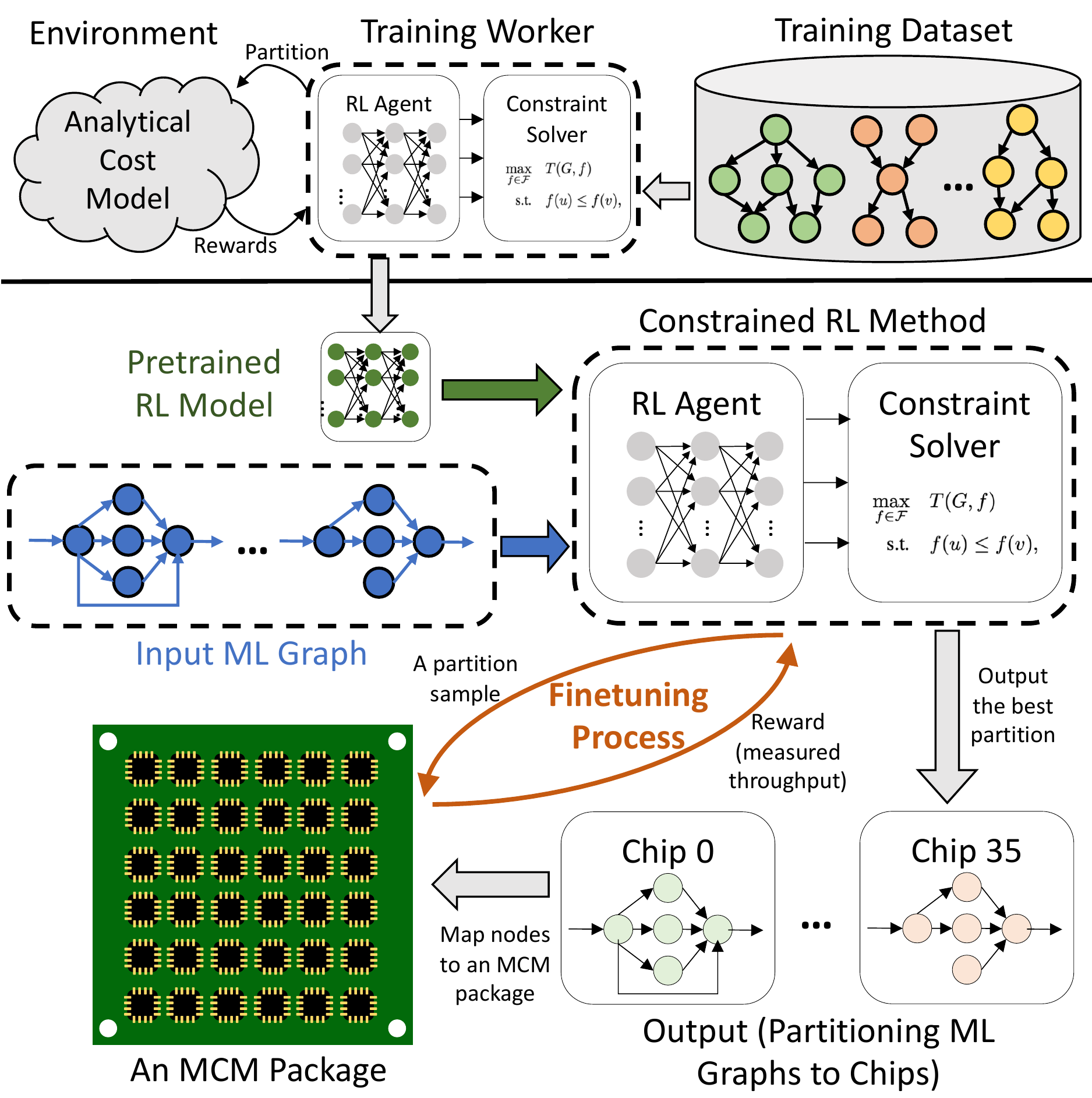}
    \caption{The overview of our constrained RL method that includes an offline pre-training on a training dataset and an online fine-tuning to generate an ML model partition. The objective is to optimize throughput or end-to-end latency targeting an MCM package.}
    \label{fig:intro_overview}
\end{figure}

In this work, we propose a partitioner that combines RL with a constraint solver to address the aforementioned shortcomings of existing solutions. 
The deep RL engine learns to interact with the dynamic compilation environment and creates an action distribution that is biased towards a balanced partition while the constraint solver takes the distribution and generates partition solutions that respect various constraints (e.g. acyclic dataflow, no skipping chip).
The overview of our RL-based method with a constraint solver is shown in Figure~\ref{fig:intro_overview}. 
To further reduce the compilation time, we develop a pre-training and fine-tuning method to generalize the pre-trained policies to unseen input graphs. 
Our RL-based partitioner shows strong generalization via the pre-training even when using an analytical cost model during the pre-training followed by fine-tuning on real hardware. 
The use of an analytical cost model saves hardware resource consumption and reduces pre-training time from a week to hours.
During the deployment of our RL-based method, this generalization significantly reduces the number of samples and shortens the compilation time caused by expensive real hardware evaluation.
We summarize the comparison between our work and prior studies in Table~\ref{tab:intro_comp}.
Our contributions are broken down in the following order:

\begin{itemize}

\item We define the problem of multi-chip partitioning for MCMs and propose a method that combines the capabilities of deep RL networks and constraint solvers to search for good partitionings in a search space where valid solutions are extremely sparse due to constraints imposed by the hardware architecture.

\item Our evaluation for BERT, a production-scale model, on real hardware demonstrates that our RL-based partitioner achieves 6.11\% and 5.85\% higher throughput than random search and simulated annealing upon its convergence.

\item We demonstrate strong transfer learning performance via a pre-training based method. 
We pre-trained the RL policy on 66 production neural networks from computer vision applications and language models, using an analytical performance model as a reward function. 
Fine-tuning the pre-trained policy on BERT reduces the search time from more than 3 hours for RL training from scratch to only 9 minutes, while achieving the same runtime performance. 
\end{itemize}

\begin{table}[!t]
    \centering
    \small 
    \begin{tabular}{C{43pt}|c|c|c|c|c}
    \hline
         & CPS & CH & RL & CPS + S & \textbf{CPS + RL} \\
         & & & & & \textbf{(Our Work)} \\
    \hline
    Static Constraints & Yes & Yes & No & Yes & \textbf{Yes} \\
    \hline
    Dynamic Constraints & No & Yes & No & Yes & \textbf{Yes} \\
    \hline
    Requires Close-Form Perf. Model & Yes & No & No & No & \textbf{No} \\
    \hline
    Solution Quality & N.A. & Low & N.A. & Medium & \textbf{High} \\
    \hline
    Time to Solution & N.A. & Fast & N.A. & Slow & \textbf{Fast} \\
    \hline 
    \end{tabular}
    \caption{Comparison among Constraint Programming Solver (CPS), Compiler Hueristic (CH), Reinforcement Learning (RL), Search Algorithms with Constraint Solvers (CPS + S), and RL with Constraint Solvers (CPS + RL).}
    \vspace{-1em}
    \label{tab:intro_comp}
\end{table}

\section{Related Work}

\textbf{Multi-Chip-Module Package:}
The advance of interconnect and package technology drives the the development of multi-chip-module (MCM) architectures to reduce the design and fabrication cost, e.g., CPU~\cite{kannan2015enabling}, GPU~\cite{arunkumar2017mcm}, and ML accelerators~\cite{zimmer20190,shao2019simba}.
Instead of building a large, monolithic chip, an MCM combines small chiplets into a chip package, thus reduces the per-chip complexity and improves manufacture yields. However, coming up with a balanced workload partitioning while minimizing inter-chip communication becomes critical for high workload throughput on an MCM package, as the off-chip communication is both low bandwidth and high latency.
Both hardware methods such as active interposers~\cite{jerger2014noc,stow2017cost} and software methods such as compiler heuristics~\cite{shao2019simba} are developed for the partitioning and mapping problem.
We are the first to develop a novel reinforcement learning method tailoring the workload partitioning and placement problem, targeting an edge TPU-based MCM.

\textbf{Model Parallelism:}
Power-law in deep learning drives the design of large and complex neural models~\cite{hestness2017deep,shazeer2017outrageously,jozefowicz2016exploring,mahajan2018exploring,radford2019language}. 
Model parallelism enables running large models on hardware devices, particularly for edge devices where on-chip memory is scarce. Mesh-TensorFlow~\cite{shazeer2018meshtensorflow}, GPipe~\cite{Gpipe2018cvpr}, GShard~\cite{lepikhin2020gshard} provide various programming primitives to enable the execution of large models on a cluster of hardware accelerators.
Pipeline parallelism~\cite{Gpipe2018cvpr, PipeDream2019sosp} enables model parallelism along the temporal axis, yet still maps computational graphs across multiple accelerators for higher throughput.
Although these studies provide computational primitives and conduct heuristic-based optimizations, none of them targets partitioning ML models on an MCM package with hardware constraints.
In this work, we develop an automatic model partitioning solution targeting an MCM package.

\textbf{ML for automatic partition and placement:}
Reinforcement learning has been used for device placement \cite{Mirhoseini2017ICML, pmlr-v80-gao18a, Mirhoseini2018ICLR, zhou2021transferable} and has demonstrated run time reduction over human-crafted placements and conventional heuristics. Progressive placements~\cite{Mirhoseini2018ICLR, pmlr-v80-gao18a, Addanki2019Placeto},  generate decisions on a per-node basis, so they have difficulty capturing long-distance dependencies over large graphs and are slow in training. 
Placeto~\cite{Addanki2019Placeto} represents the first attempt to generalize device placement using a graph embedding network. 
GO~\cite{zhou2021transferable} is the first single-shot method that generates placement decisions for an entire graph and generalizes to unseen data. 
However, all the above methods do not handle constraints explicitly and can fail when facing strict systems constraints imposed by novel architectures because of an ultra-sparse reward space. 

\textbf{Constrained learning:} 
Differentiable SAT~\cite{wang2019satnet} and differentiable optimizer~\cite{amos2019optnet} integrate constraints to the end-to-end learning systems, which can learn the logical structure via supervised learning. 
Constrained Policy Gradient~\cite{achiam2017constrained} trains neural network policies for high-dimensional control while making guarantees about all policy behaviors throughout training. 
However, many system constraints can hardly be formulated statically during the compilation, necessitating an actor to interact with the environment and learn the interacting constraints.
None of these studies are tailored for finding optimal solutions in a compilation system with dynamic constraints because of the lack of a closed-form objective formulation. 
In this work, we leverage the constraint solvers to rule out infeasible partition statically while using reinforcement learning to propose optimal solutions through interaction with a real system.
\section{Hardware Architecture and Problem Formulation}

Hardware specialization provides speedups and higher energy efficiency for ML workloads.
However, some specialized hardware impose constraints when mapping the workload during the compilation stage. 
In this section we introduce the target hardware platform of our problem--a multi-chip TPU~\cite{dasari2021apparatus}, then define and formulate the multi-chip placement problem with constraints.

\noindent \textbf{Hardware Architecture:}
In this work, our target hardware platform is a 36-die multi-chip ML accelerator~\cite{dasari2021apparatus} package with a 1D ring for inter-chip communication. 
Figure~\ref{fig:placement_problem}b shows an example 4-die multi-chip package with only uni-directional links among adjacent chips, and the inter-chip link topology of our multi-chip TPU package is similar to that of Figure~\ref{fig:placement_problem}.
Each chip has tens of MBs SRAM, and inter-chip links only offer a bandwidth of tens of GB/s.
Thus, it is unavoidable to partition production-scale ML models to our multi-chip TPU, and optimizing inter-chip communication is important.

\noindent \textbf{Problem Definition:} 
Denote the set of available chips as $D=\lbrace 0, 1, 2, ..., C - 1 \rbrace$, and a directed acyclic graph $G=(V, E)$ representing an ML workload where $V$ stands for the vertex set of operations and $E$ stands for the edge set of dependencies between operations.

The multi-chip partitioning problem aims to find a function $f$, which maps $V$ to $D$ denoted as $f: V \mapsto D $, that maximizes or minimizes an objective function.
Figure~\ref{fig:placement_problem}a shows an example of a computation graph for an ML workload.
In this work, we aim to maximize the throughput of ML workloads on ML accelerators. 
Thus the multi-chip partitioning task can be formulated as Equation~\ref{eq:placement_task} denoting $T(G, f)$ as the throughput of running the graph $G$ on the hardware with the mapping function $f$, and $\mathcal{F}$ as the set of all possible mapping functions from $V$ to $D$.

\begin{equation}
\begin{aligned}
\max_{f \in \mathcal{F}} \quad & T(G, f) \quad &
\end{aligned}
\label{eq:placement_task}
\end{equation}

\begin{figure*}
    \centering
     \includegraphics[width=0.82\textwidth]{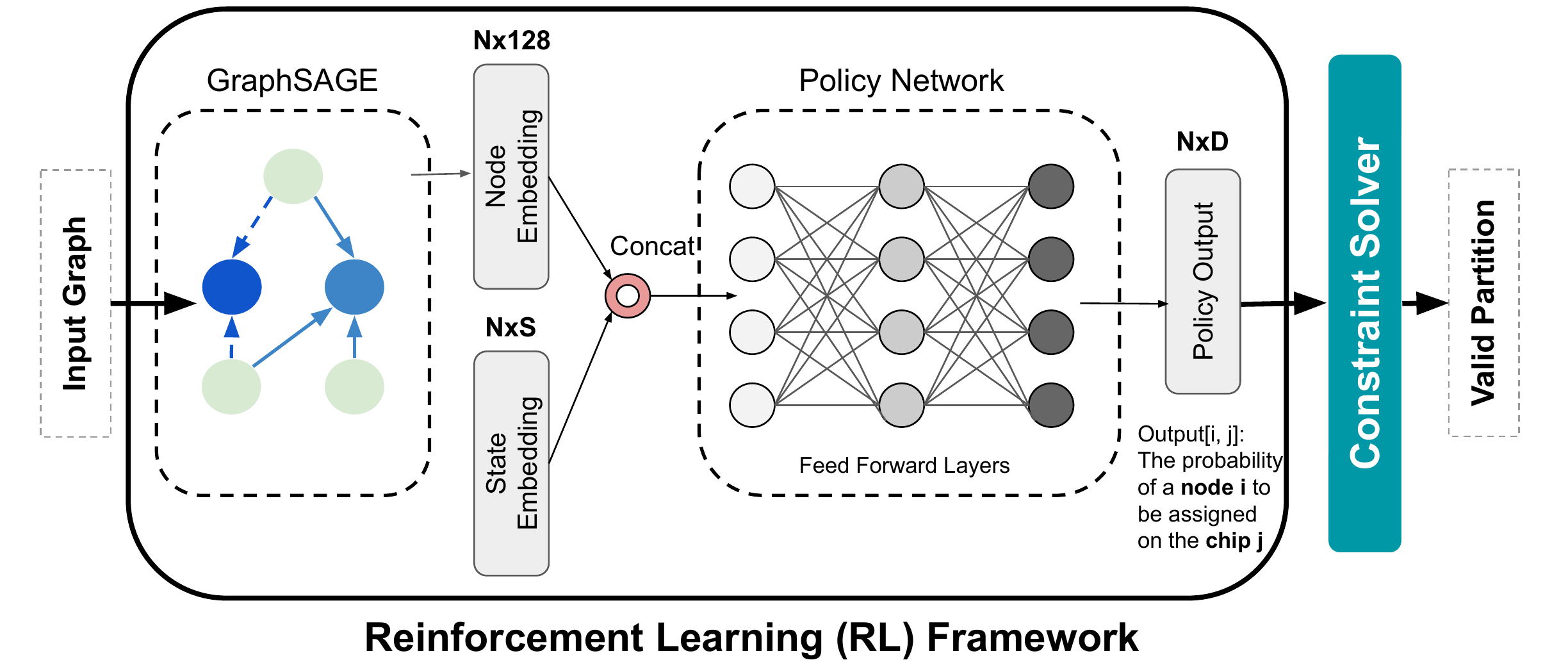}
    \caption{The overview of our RL-based method including an RL framework to generate the probability distribution of chip assignments, and a constraint solver to generate a valid partition by sampling according to the policy output from the RL.}
    \label{fig:go_telamon_overview}
\end{figure*}

\noindent \textbf{\emph{Constraint 1: Acyclic Dataflow Constraint.}}
Since the links among chips are uni-directional, only data transfer from low chip ID from high chip ID is allowed.
For example, Figure~\ref{fig:placement_problem}c shows an invalid partition where data transfer from chip 1 to chip 0 between node 2 and node 4 is not allowed.
Denoting $f(u)$ as the chip ID where the node $u$ is mapped to using the mapping function $f$, this constraint can be formulated as Equation~\ref{eq:dataflow_constraint}.
\begin{equation}
\begin{aligned}
f(u) \le f(v) \quad & \forall (u, v) \in E \\
\end{aligned}
\label{eq:dataflow_constraint}
\end{equation}

\noindent \textbf{\emph{Constraint 2: No Skipping Chips.}}
Since the ML accelerator can pipeline the execution of operators mapped to chips, the backend of the current multi-chip TPU compiler does not allow skipping chips to maximize the throughput. 
As a part of compilation optimization, we also impose this constraint on this multi-chip partitioning task.
For example, Figure~\ref{fig:placement_problem}d shows an invalid partition where the chip 1 is skipped without any nodes mapped to it.
This constraint can be formulated as Equation~\ref{eq:no_skipping_devices}.
\begin{equation}
\begin{aligned}
d \le f(u) \Rightarrow \exists v \in V, f(v) = d \quad & \forall u \in V, d \in D\\
\end{aligned}
\label{eq:no_skipping_devices}
\end{equation}

\noindent \textbf{\emph{Constraint 3: Chip Triangle Dependency.}}
Because the simplified network-on-chip (NoC) routers on the multi-die TPU cannot handle all NoC traffic patterns, direct data dependencies between two chips cannot co-exist with indirect data dependencies between the same chips. 
For example, Figure~\ref{fig:placement_problem}e shows an invalid partition where there is a direct dependency between chip 0 and chip 2 through the data transfer between node 0 and node 2 while there is an indirect dependency chain from chip 0 to chip 1 then chip 2 through the data transfer from node 0 to node 1 then node 3.
To formulate this dependency, we define $\delta(d_0, d_1)$ as the length of the longest path from chip $d_0 \in D$ to chip $d_1 \in D$ in the graph $G$ whose nodes are the chips and edges are data dependencies between chips. We then impose this length to be at most one for each direct dependency in the computational graph, as formulated in Equation~\ref{eq:chip_dependency}.
\vspace{-1em}

\begin{equation}
\begin{aligned}
\delta(d_0, d_2) \geq \delta(d_0, d_1) + \delta(d_1, d_2) \quad & \forall d_0, d_1, d_2 \in D \\
f(u) \neq f(v) \Rightarrow \delta(f(u), f(v)) = 1 \quad & \forall (u, v) \in E \\
\end{aligned}
\label{eq:chip_dependency}
\end{equation}

\noindent \textbf{\emph{Constraint 4: Dynamic Constraint.}}
In addition to static constraints which can be explicitly expressed in closed-form formulas, there are also constraints from system dynamics. 
For example, the on-chip memory consumption of a model partition (mapping $f$) should fit in memory.
Figure~\ref{fig:placement_problem}f shows an invalid partition that leads to the out-of-memory allocation issue by consuming too much on-chip memory on chip~1.
Because these dynamic constraints are not able to be explicitly formulated, we define a boolean function $H(G, f)$ which returns true only if the mapping function $f$ can pass the compilation and hardware evaluation.
To ensure the completeness of the problem formulation, we add this boolean function as a constraint in the overall formulas.

\noindent \textbf{Putting It All Together.}
Combining all static constraints from the hardware architecture of multi-chip TPU and dynamic constraints from the compiler backend, the multi-chip partitioning task that maps a computation graph $G$ to the available set of chips $D$ can be formulated as Equation~\ref{eq:placement_problem}.

\begin{equation}
\footnotesize
\begin{aligned}
\max_{f \in \mathcal{F}} \quad & T(G, f) \quad & \\
\textrm{s.t.} \quad & f(u) \le f(v) \quad & \forall (u, v) \in E \\
            \quad & d \le f(u) \Rightarrow \exists v \in V, f(v) = d \quad & \forall u \in V, d \in D \\
            \quad & \delta(d_0, d_2) \geq \delta(d_0, d_1) + \delta(d_1, d_2) \quad & \forall d_0, d_1, d_2 \in D \\
            \quad & f(u) \neq f(v) \Rightarrow \delta(f(u), f(v)) = 1 \quad & \forall (u, v) \in E \\
            \quad & H(G, f) = True \\
\end{aligned}
\label{eq:placement_problem}
\end{equation}

Although prior studies develop reinforcement learning solutions, strict constraints in Equation~\ref{eq:placement_problem} invalidate most of the possible mappings from the set $\mathcal{F}$.
Thus, the reward space is ultra-sparse, making it difficult for the agent to learn an optimal partition.
On the other hand, although constraint solvers are good at solving optimization problems, they are not able to handle non-closed form functions. 
Because the objective function $T(G, f)$ and the dynamic constraint $H(G, f)$ cannot be explicitly formulated, it is impossible to apply constraint solvers directly for solving the problem formulated in Equation~\ref{eq:placement_problem}.
These shortcomings motivate us to develop our constrained reinforcement learning method detailed in Section~\ref{sec:constrained_rl_solution}, which combines the reinforcement learning method with the constraint solver to efficiently explore the space of feasible partitions.

\section{Reinforcement Learning with a Constraint Solver} \label{sec:constrained_rl_solution}

The overview of our RL-based approach is shown in Figure~\ref{fig:go_telamon_overview}. 
In this section, we will detail our RL framework for generating a chip assignment for nodes in Section~\ref{sec:rl_solution}, and the constraint solver for generating a valid ML model partition in Section~\ref{sec:telamon}. 
Then, we will introduce our pre-training pipeline to generalize our RL-based approach in Section~\ref{sec:pretraining}.

\subsection{Reinforcement Learning} \label{sec:rl_solution}

As shown in Figure~\ref{fig:go_telamon_overview}, our learnt policy $\pi_\theta$ (policy network) takes the node embedding of a graph, $h_{G}$, from a feature network and the current state embedding and generates a probability distribution matrix $\mathbf{P} = [\mathbf{p}_{1}, \mathbf{p}_{2}, ..., \mathbf{p}_{N}]$ where $N$ stands for the number of nodes and the vector $\mathbf{p}_{i} = [p_{i1}, p_{i2}, ..., p_{iC}]$ stands for the probability distribution of $i$-th node to $C$ available chips.
We adopt GraphSAGE~\cite{William2017NIPS} for the feature extraction of an input computation graph to generate $h_{G}$, and a fully connected network as the policy network.
We train the feature extraction network and the policy network in an end-to-end fashion, using a reward of the execution throughput on the target multi-chip environment.

Let $y_i$ be the action for the $i$-th node ($y_i \in D$). 
Ideally, we would like to compute the action distribution of the current node based on the actions of all previous nodes in an auto-regressive manner:
\begin{equation}
\label{equation:formulation}
    p(\mathbf{y}|G) = \prod^{N}_{i=1} p\left(y_i | h_G, y_{i-1}, y_{i-2}, ...\right)
\end{equation}
However, the above is infeasible in practice because the number of nodes can be as large as 10K, and computing the $y_i$'s sequentially can be extremely expensive.
To address this issue, we use an iterative but non-autoregressive process as an approximation same as prior work~\cite{zhou2021transferable}:
\begin{equation}
    p(\mathbf{y}^{(t)}|G) = \prod^{N}_{i=1} p\left(y^{(t)}_i | h_G, \mathbf{y}^{(t-1)}\right)
\end{equation}
Although the $N$ sampling procedures are now carried out in parallel within each iteration $t$, decisions over the $N$ nodes are allowed to mutually influence each other because the process above will be repeated for $T$ times ($T \ll N$). 
Note the distribution of $\mathbf{y}^{(t)}$ is informed about $\mathbf{y}^{(t-1)}$, the actions made over all the nodes in the previous iteration.
At each iteration, a concrete partition solution $\mathbf{y}^{(t)}$ can be sampled from the probability distribution $\mathbf{P}^{(t)}$.

Due to constraints formulated in Equation~\ref{eq:placement_problem}, the reward space for the action $\mathbf{y}$ is extremely-sparse.
Thus we use the reward of $\mathbf{y'}$ rather than directly using the reward of $\mathbf{y}$ to efficiently learn policy $\pi_\theta$, where $\mathbf{y'}$ is a valid partition generated from the constraint solver according to $\mathbf{y}$ and partitioning problem constraints. 
The constraint solver is detailed in Section~\ref{sec:telamon}.

\subsection{Constraint Solver} \label{sec:telamon}

The role of the constraint solver in Figure~\ref{fig:go_telamon_overview} is to find a valid partition of the constraint satisfaction problem that follows problem constraints. 

The procedure to find a solution is based on the CP-SAT open-source constraint solver~\cite{CP-SAT}.
The solver has a variable representing action $y_i$ for each node $i$ and enforces constraints in equations (\ref{eq:dataflow_constraint})-(\ref{eq:chip_dependency}).
Internally, the solver maintains the range of valid values for every variable $y_i$, called the \emph{domain} of $y_i$.
The procedure to find a solution interacts with the solver by querying the current domain of variables and by manually setting variable domains.
When setting the domain of a variable $y_i$, the solver internally runs a \emph{constraint propagation} algorithm that recursively prunes the domain of other variables so that they only contain values that are compatible with the new domain of $y_i$ with regards to constraints.
If constraint propagation detects an invalid assignment, the solver will backtrack to a previous state, undoing previous decisions.

Our procedure finds a solution by picking the assignment of one node at a time.
At each step, it picks a node $i$ and queries the current domain $\mathcal{D}_i$ of $y_i$ from the solver.
It then calls the solver to restrict the domain of $i$ to a single value $c \in \mathcal{D}_i$ taken from the domain of $y_i$.
The procedure stops when all nodes are assigned a device.

\begin{algorithm}[!t]
\begin{algorithmic}
\caption{Constraint solver in SAMPLE mode}
\label{algm:sample}
\footnotesize
\STATE \textbf{Input}: A node order $\mathcal{T}$, and a distribution $\mathbf{P}$.
\STATE \textbf{Output}: A valid partition $\mathbf{y'}$.
\STATE $S$ = init\_solver()  \codecomment{// Init the constraint solver}
\STATE $i = 0$
\WHILE{$i < N$}
    \STATE $u = \mathcal{T}_{i}$
    \STATE \codecomment{// Get the current valid domain of the node $u$.}
    \STATE $\mathcal{D}_u$ = $S$.get\_domain($u$)
    \STATE $y'_{u}$ = sample $\mathbf{p}_{u}$ from $\mathcal{D}_{u}$
    \STATE \codecomment{// Set $y_u$ domain, perform constraint propagation, and backtrack to a previous $i$ if needed}
    \STATE $i$ = $S$.set\_domain($u$, $\{y'_{u}\}$)
\ENDWHILE
\end{algorithmic}
\end{algorithm}

\begin{algorithm}[!t]
\begin{algorithmic}
\caption{Constraint solver in FIX mode}
\label{algm:fix}
\footnotesize
\STATE \textbf{Input}: A node order $\mathcal{T}$, and a candidate partition $\mathbf{y}$.
\STATE \textbf{Output}: A valid partition $\mathbf{y}'$.
\STATE $S$ = init\_solver()  // Init the constraint solver
\STATE $i = 0$
\WHILE{i < 2 $\times$ N}
    \STATE $u = \mathcal{T}_{i \text{ mod } N}$
    \IF{i < N}
        \STATE $\mathcal{D}_u$ = $S$.get\_domain($u$)
        \IF{$y_u \in \mathcal{D}_u$}
            \STATE $y'_{u} = y_u$ 
            \STATE $i$ = $S$.set\_domain($u$, $\{y'_u\}$)
        \ELSE
            \STATE $i$ = $S$.set\_domain($u$, $\mathcal{D}_u$)
        \ENDIF
    \ELSE
        \STATE $\mathcal{D}_u$ = $S$.get\_domain($u$)
        \STATE $y'_u = $ randomly sample $\mathcal{D}_{u}$
        \STATE $i$ = $S$.set\_domain($u$, $\{y'_{u}\}$)
    \ENDIF
\ENDWHILE
\end{algorithmic}
\end{algorithm}

There are two important factors for our constraint solver to generate a valid partition $\mathbf{y'}$ given an input partition $\mathbf{y}$ and a probability distribution $\mathbf{P}$:
(1) the node traversal order, and (2) the strategy to pick a value $y_i$ from the valid domain.
First, our constraint solver provides an interface to specify the node order.
By default, we generate a random order each time to explore a larger decision space rather than prioritizing a fixed set of nodes that significantly prunes the domain of other nodes. 
Second, there are two strategies to pick a value $y_i$ from the current valid domain.

\begin{itemize}
    \item With the \textsc{SAMPLE} strategy, the solver visits nodes according to the input node order. For each node $i$, it samples a chip ID according to $\mathbf{p}_{i}$ restricted to the current domain of the $y_i$.

    \item With the \textsc{FIX} strategy, the constraint solver traverses nodes according to the input node order, and assigns $y_i$ to $y'_i$ if $y_i$ is a valid assignment. After the traversal, it repeatedly assigns random chip IDs to remaining nodes (where $y_i$ is invalid) until getting a valid assignment. 
\end{itemize}

The algorithms for SAMPLE and FIX strategy are detailed in Algorithm~\ref{algm:sample} and Algorithm~\ref{algm:fix} respectively.
In these two algorithms, we rely on the underlying CP-SAT solver (denoted as $S$) to maintain and update valid domains for all nodes.
In particular, the function \textit{get\_domain} returns the current valid domain for a node, and the function \textit{set\_domain} performs the constraint propagation step to update valid domains for all nodes.
The loop index, $i$, is the number of decisions set by calling \textit{set\_domain}.
Calling \textit{set\_domain} returns the new value of $i$.
In general, this is $i+1$ but this can also be a lower value if the solver backtracks.

As shown in Figure~\ref{fig:go_telamon_overview}, RL model outputs both $\mathbf{y}$ and $\mathbf{P}$ so that the constraint solver generates a valid partition $\mathbf{y'}$ according to the input node order and assignment strategy. 
Because $\mathbf{y'}$ satisfies static constraints, its reward space is denser than that of $\mathbf{y}$, and we use this denser reward space to efficiently train our end-to-end RL models.

\subsection{Pretraining Pipeline} \label{sec:pretraining}
\begin{figure*}
    \centering
     \includegraphics[width=0.98\textwidth]{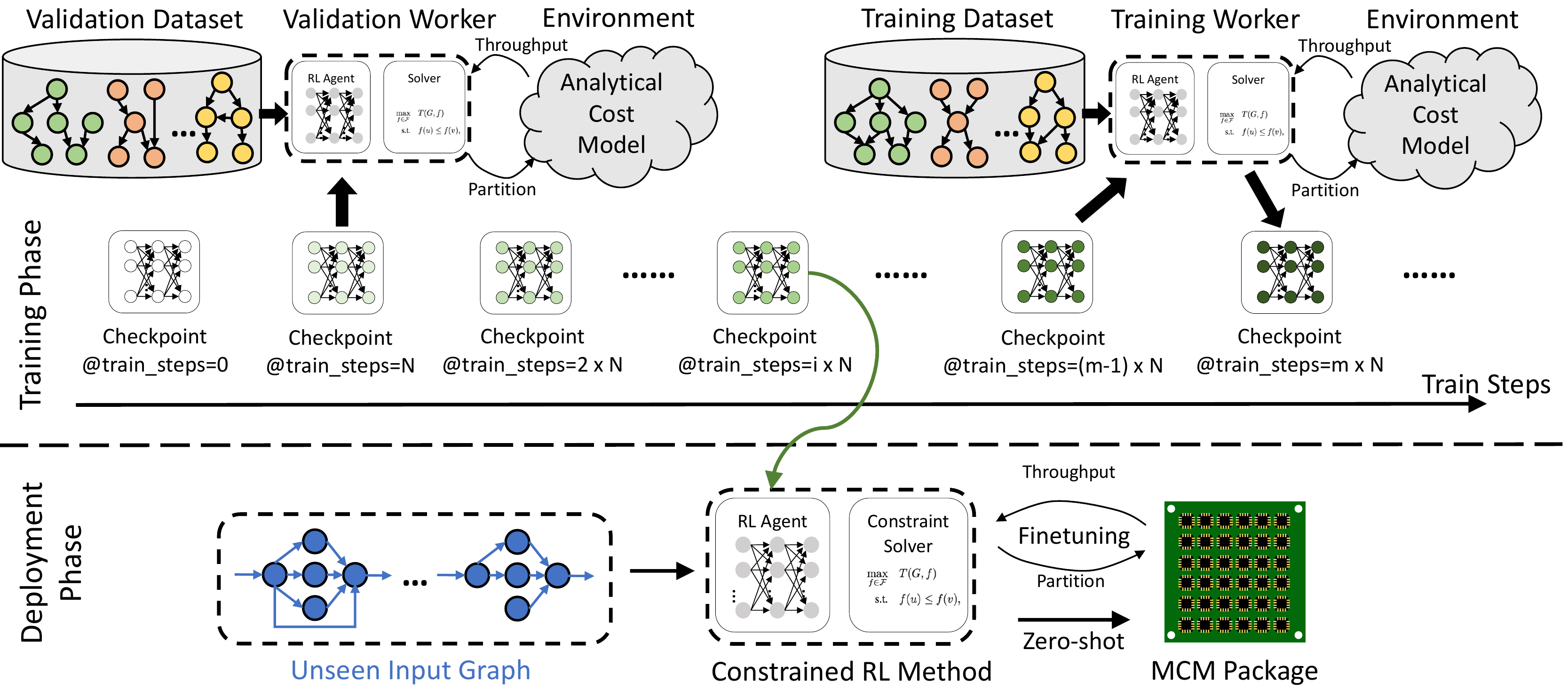}
    \caption{The workflow of our pre-training method including a training phase to obtain pre-trained model checkpoints, and a deployment phase using the optimal checkpoint for zero-shot and fine-tuning on an unseen input graph.}
    \label{fig:pretraining}
\end{figure*}
In addition to the constrained deep RL method introduced in Section~\ref{sec:rl_solution}, we develop a pre-training based method to generalize the constrained deep RL solution such that it can learn from the training dataset while transferring the learned knowledge to unseen data.
As shown in Figure~\ref{fig:pretraining}, the whole pre-training pipeline is composed of two phases: the training phase and the deployment phase.

\noindent \textbf{Training Phase:}
There are two workers in the training phase: a training worker generating the checkpoints of RL model weights, and a validation worker to evaluate the performance of each model snapshot.
In particular, the training worker iterates through the input graphs from the training set, and periodically generates checkpoints of the RL model weights.
Meanwhile, the validation worker conducts a continuous evaluation on graphs from the validation set. The validation worker is warm-started from a pre-trained model checkpoint. 
During the evaluation, the validation worker conducts a zero-shot prediction and a fine-tuning upon the pre-trained checkpoint, for each graph from the validation set. 
After iterating through all model checkpoints, the validation worker can pick the checkpoint with either the best zero-shot or fine-tuning performance for deployment. 

\noindent \textbf{Deployment Phase:}
In the deployment phase, we load the optimal checkpoint picked by the validation worker to warm start the RL model.
Then the RL model takes a new (previously unseen) graph as the input and directly runs inference (zero-shot). 
Alternatively, we can fine-tune the RL model for this new graph to capture out-of-distribution data, as detailed in Figure~\ref{fig:go_telamon_overview}.

Experimental results in Section~\ref{sec:overall_perf} show that our pre-training pipeline is able to generalize the pre-trained solution to a test set of previously unseen graphs.
\section{Experiments} \label{sec:experiment}

Our goal is to find a valid and efficient ML model partition to chips while minimizing the search time.
First, we introduce the experiment setup in Section~\ref{sec:exp_setup}. 
Then, we present the results of our pre-training experiments in Section~\ref{sec:overall_perf} to demonstrate the generalization of our RL approach on the test dataset of 16 ML graphs.
Finally, we evaluate the BERT model in Section~\ref{sec:bert_studies} to show the effectiveness of our RL approach on real hardware.  

\subsection{Experiment Setup} \label{sec:exp_setup}

\textbf{Evaluation Platform:}
We conduct our experiments on a real hardware platform of multi-chip TPU~\cite{dasari2021apparatus} that includes 36 chiplets in a package. 
To efficiently pre-train our RL model, reducing pre-training time, and saving hardware resources, our pre-training experiments are based on an analytical cost model.
This analytical cost model estimates the latency of running all nodes assigned to each chip, and returns the maximal latency of all chips.

\textbf{Workloads:}
We conduct the real hardware evaluation on a production-scale model, BERT~\cite{BERT}, that has 2138 nodes and around 340 million (600 MB) parameters.
To pre-train our RL model, we use a dataset of 87 ML models from real-world applications, and most of them are from computer vision and natural language processing applications including convolutional neural network (CNN) and recurrent neural network (RNN) models.
The computation graphs of these ML models have tens to hundreds of nodes.
None of these ML graphs contain a Transformer-like attention mechanism.
In our pre-training experiment, we randomly partition these 87 computation graphs into three datasets: a training dataset of 66 graphs, a validation set of 5 graphs, and a test dataset of 16 graphs.

\textbf{Performance Metric:}
We evaluate each partition solution and obtain its throughput because a multi-chip TPU focuses more on throughput rather than latency. 
However, our framework can easily re-target a latency metric. 
Then we report the throughput improvement over compiler heuristics, such as a greedy algorithm and a random partition, that are usually fast with a time complexity of $O(N)$.
We run all experiments on real hardware 5 times and report both mean and standard deviation of throughput improvements. 

\textbf{RL with Constraint Solver:}
Our default configuration of graph neural network uses 8 GraphSAGE layers, and the size of each layer is 128.
For the policy network, we use a feed-forward network with 2 layers, and the size of each layer is the same as GraphSAGE layers.
We use Proximal Policy Optimization (PPO)~\cite{PPO2017} to train our RL models.
We explore various values for hyper-parameters during the training process, such as the number of rollouts, the number of mini-batches, and the number of epochs. 
We select the optimal hyper-parameter (20 rollouts, 4 mini-batches, and 10 epochs) across all explored settings to report RL results.
For this example, we use the FIX mode in the constraint solver, as it outperforms SAMPLE mode.

\textbf{RL without Constraint Solver:} 
Instead of passing the policy output from RL to the constraint solver, this baseline directly generates partition solutions by sampling based on the output probability distribution $\mathbf{P}$.
Our evaluation platform returns a zero throughput when it evaluates an invalid partition.
Without the help of the constraint solver, the reward space is extremely sparse as most of the partitions are invalid.
As a result, this baseline is not able to find any valid partition in the training process even when the number of samples is sufficiently large. 

\textbf{Traditional Search Strategies:}
We implement several traditional search strategies working with the constraint solver as comparison baselines. 
These traditional search strategies include \emph{Random Search (Random)} and \emph{Simulated Annealing (SA)}.
We tune these search strategies with the constraint solver empirically to pick an implementation with the best performance. 
The details of each search algorithm are as follows:

\emph{Random Search Strategy (Random)}
provides a fixed uniform probability distribution, $p(y_i=j|G) = \frac{1}{C}$ for $\forall j \in D$, and the constraint solver works under the SAMPLE mode to generate valid partitions.   

\emph{Simulated Annealing (SA)}
starts from the same uniform probability distribution as Random. 
Each iteration, SA randomly selects a set of nodes, and for each node $i$ from this set, it generates a new random distribution $\mathbf{p}_{i}$.
The constraint solver takes this new probability distribution to generate a valid partition. 
Based on the evaluation results of this valid partition, SA decides whether to accept the new probability distribution.

\subsection{Pretraining Experiment} \label{sec:overall_perf}

\begin{figure}[!t]
    \centering
    \includegraphics[width=1.0\linewidth]{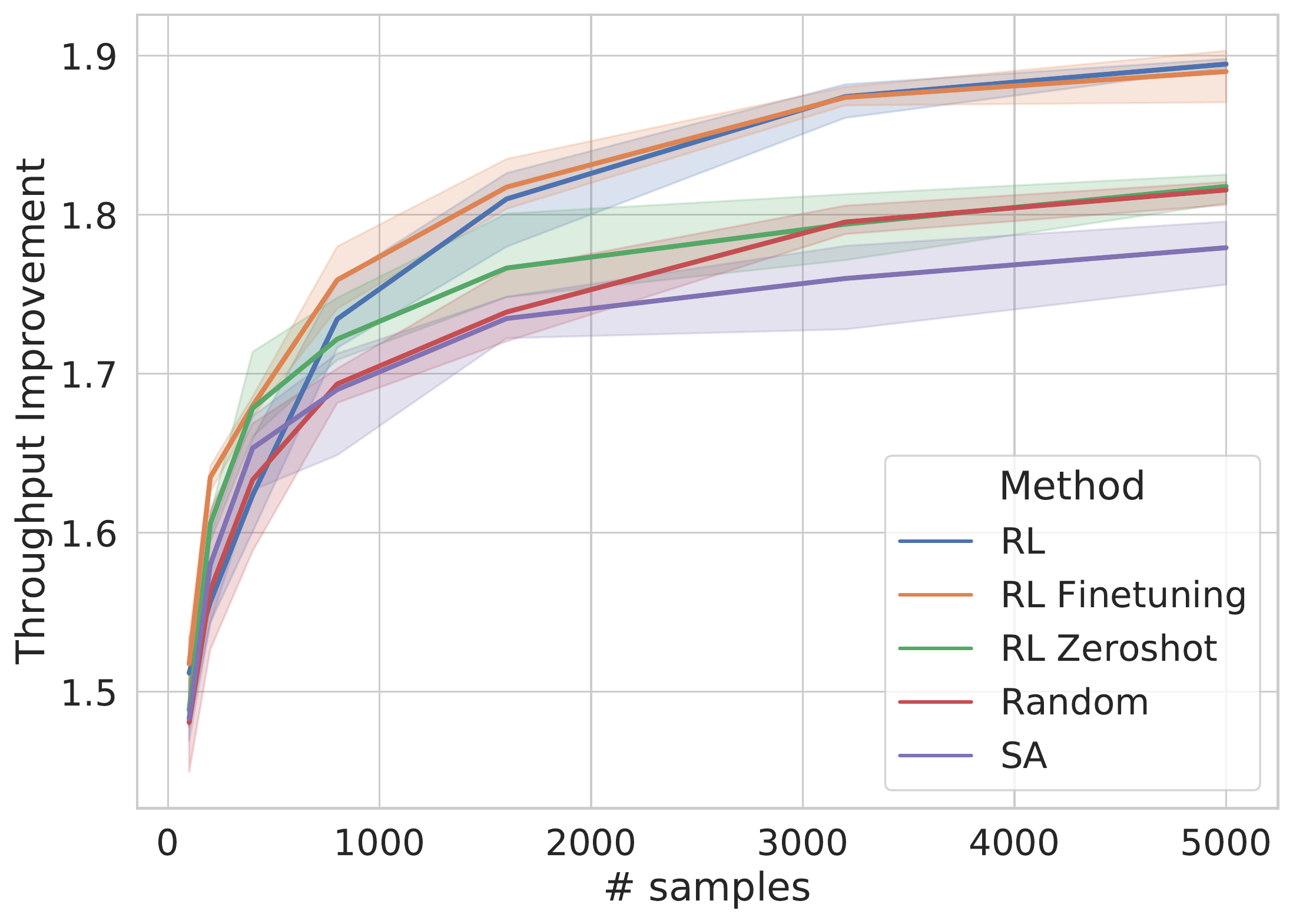}
    \caption{The geomean throughput improvement of 16 graphs from test dataset on the analytical model comparison for a random search strategy (Random), simulated annealing (SA), reinforcement learning training from scratch (RL), the zeroshot predictions of the pretrained model (RL Zeroshot), and the finetuning of the pretrained model (RL Finetuning).}
    \label{fig:perf_overall}
\end{figure}

\begin{table}[!t]
    \centering
    \small 
    \begin{tabular}{L{52pt}|rrr}
	    \hline
	    Throughput Improvement & $\ge$ 1.60$\times$ & $\ge$ 1.70$\times$ & $\ge$ 1.80$\times$ \\
	    \hline
	    Random  & 305 (1.08$\times$) & 915 (0.74$\times$) & 3612 (0.41$\times$)\\ 
	    SA & 255 (1.29$\times$) & 979 (0.69$\times$) & N.A. (N.A.) \\
	    RL & 330 (1.00$\times$) & 676 (1.00$\times$) & 1496 (1.00$\times$) \\
	    RL Zeroshot & 196 (1.68$\times$) & 600 (1.13$\times$) & 3652 (0.41$\times$) \\
	    RL Finetuning & \textbf{171 (1.93$\times$)} & \textbf{503 (1.34$\times$)} & \textbf{1362 (1.10$\times$)} \\
	    \hline
    \end{tabular}
    \caption{The number of samples and the reduction of samples to achieve certain geomean throughput improvement levels. The results 171 (1.93$\times$) indicate that RL Finetuning needs 171 samples and reduces 1.93$\times$ samples compared with RL training from scratch to achieve a 1.60$\times$ geomean throughput improvement over a compiler heuristic.}
    \vspace{-1.2em}

    \label{tab:sample_efficiency}
\end{table}

We pre-train the RL model using evaluations from an analytical model. The evaluation with an analytical model is orders of magnitude faster than evaluating the samples on real chips. 
In particular, we pre-train the RL model on the training dataset (66 graphs) with a total of 20,000 samples. The pre-training process generates 200 checkpoints. 
Then, we use the validation dataset (5 graphs) to pick the optimal checkpoint with the best average rewards.
The pre-training process takes a few hours using the analytical cost model rather than several days compared to using real-chip evaluations.
Finally, we evaluate 16 graphs from the test dataset on the analytical model to demonstrate the generalization.

\begin{figure}[!t]
    \centering
    \includegraphics[width=1.0\linewidth]{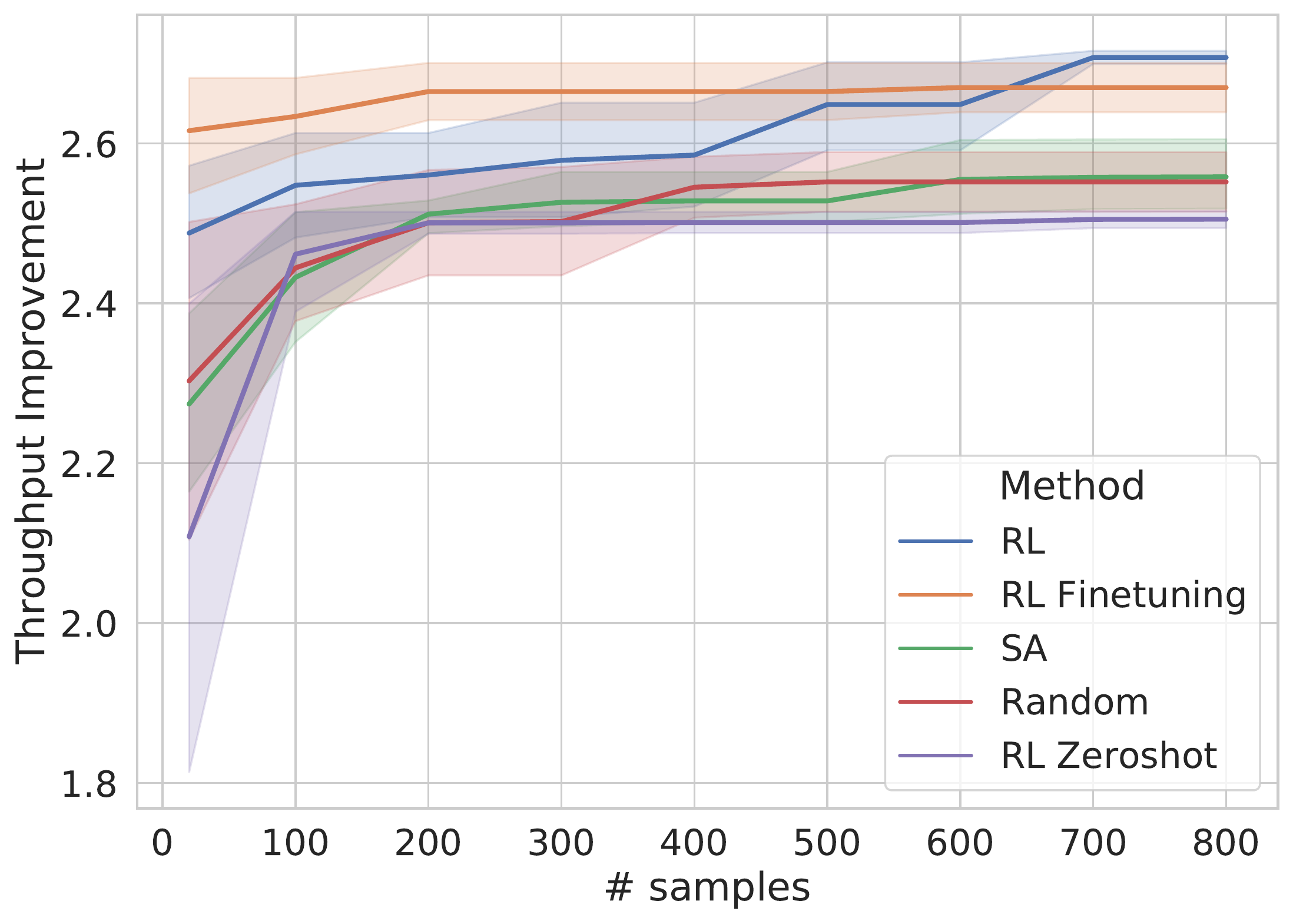}
    \caption{The throughput improvement of BERT on real hardware over a greedy heuristic comparing the random search strategy (Random), simulated annealing (SA), reinforcement learning training from scratch (RL), the zeroshot of the pretrained model (RL Zeroshot), and the finetuning of the pretrained model (RL Finetuning).}
    \label{fig:bert_perf_overall}
\end{figure}

\begin{table}[!t]
    \centering
    \small
    \begin{tabular}{L{52pt}|rrr}
	    \hline
	    Throughput Improvement & $\ge$ 2.55$\times$ & $\ge$ 2.60$\times$ & $\ge$ 2.65$\times$ \\
	    \hline
	    Random  & 447 (0.25$\times$) & N.A. (N.A.) & N.A. (N.A.)\\ 
	    SA & 576 (0.19$\times$) & N.A. (N.A.) & N.A. (N.A.) \\
	    RL & 112 (1.00$\times$) & 423 (1.00$\times$) & 607 (1.00$\times$) \\
	    RL Zeroshot & N.A. (N.A.) & N.A. (N.A.) & N.A. (N.A.) \\
	    RL Finetuning & \textbf{20 (5.60$\times$)} & \textbf{20 (21.15$\times$)} & \textbf{161 (3.77$\times$)} \\
	    \hline
    \end{tabular}
    \caption{The number of samples and the reduction of samples to achieve certain throughput improvement levels. The results 20 (5.60$\times$) indicate that RL Finetuning needs 20 samples and reduces 5.60$\times$ samples compared with RL training from scratch to achieve a 2.55$\times$ throughput improvement over a greedy heuristic.}
    \vspace{-1em}

    \label{tab:bert_sample_efficiency}
\end{table}

Figure~\ref{fig:perf_overall} presents the geomean throughput improvement over a compiler heuristic of 16 test graphs among RL, RL Finetuning, RL Zeroshot, Random, and SA.
An RL-based approach can achieve a better performance by 4.36\% and 6.49\% compared to Random and SA.
Table~\ref{tab:sample_efficiency} shows the number of samples needed to achieve different throughput gains.
It shows that fine-tuning on a pre-trained RL model can reduce the number of samples by up to 1.93x compared to RL training from scratch.
Both Figure~\ref{fig:perf_overall} and Table~\ref{tab:sample_efficiency} show that zero-shot RL (without fine-tuning) can achieve higher throughput in early hundred samples, thus 1.68x fewer samples than RL training from scratch to achieve a geomean of 1.60x speedup, but RL zero-shot does not perform well with more samples.
This could result from the different data distributions between the training dataset and the test dataset. 
It justifies a further fine-tuning on a pre-trained model for out-of-distribution data.

We further evaluate BERT on real chips in Section~\ref{sec:bert_studies} to demonstrate a real-world use case.

\subsection{BERT Evaluation} \label{sec:bert_studies}
We evaluate a production-scale model, BERT~\cite{BERT}, on a real MCM system with 36 chips to demonstrate real system performance. 
We use a greedy heuristic from the production compiler as the baseline of throughput improvement. 
Section~\ref{sec:exp_setup} details the implementation of our RL method and other traditional search strategies.
To generate sufficient pre-training samples, we adopt the analytical cost model in the pre-training phase, as described in Section~\ref{sec:overall_perf}.

Figure~\ref{fig:bert_perf_overall} shows that our RL-based approach can achieve 6.11\% and 5.85\% better throughput than Random and SA respectively, at convergence.
Moreover, Figure~\ref{fig:bert_perf_overall} shows that fine-tuning on a pre-trained RL model improves the placement throughput at low sample complexity, compared to RL training from scratch. 
Unfortunately, RL zero-shot does not work well due to two possible reasons: 1) BERT model is much bigger than the graphs in the training dataset and has a drastically different model architecture compared to models in the training dataset. 2) The difference between the analytical cost model and real-chip evaluations exacerbates the RL environment gap between the pre-training stage and the deployment (test) stage.

Table~\ref{tab:bert_sample_efficiency} shows that fine-tuning on a pre-trained RL model reduces the number of samples by up to 21.15x for achieving the same throughput gain compared to RL training from scratch.
Since the elapsed time of getting a sample takes 26.97 seconds on average, reducing the number of samples from 423 to 20 means a reduction of searching time from more than 3 hours to around 9 minutes.
Finally, for a search budget of 10 minutes, the fine-tuning of our RL-based approach can outperform Random and SA by up to 15.18\%.

The results demonstrate strong generalization from an analytical cost model based pre-training on training dataset consisting of small production models, to a real-chip environment on large production models such as BERT.
The pre-training and fine-tuning method enables the deployment of an RL-based method into production ML compilers.

\begin{figure}[!t]
    \centering
    \includegraphics[width=1.0\linewidth]{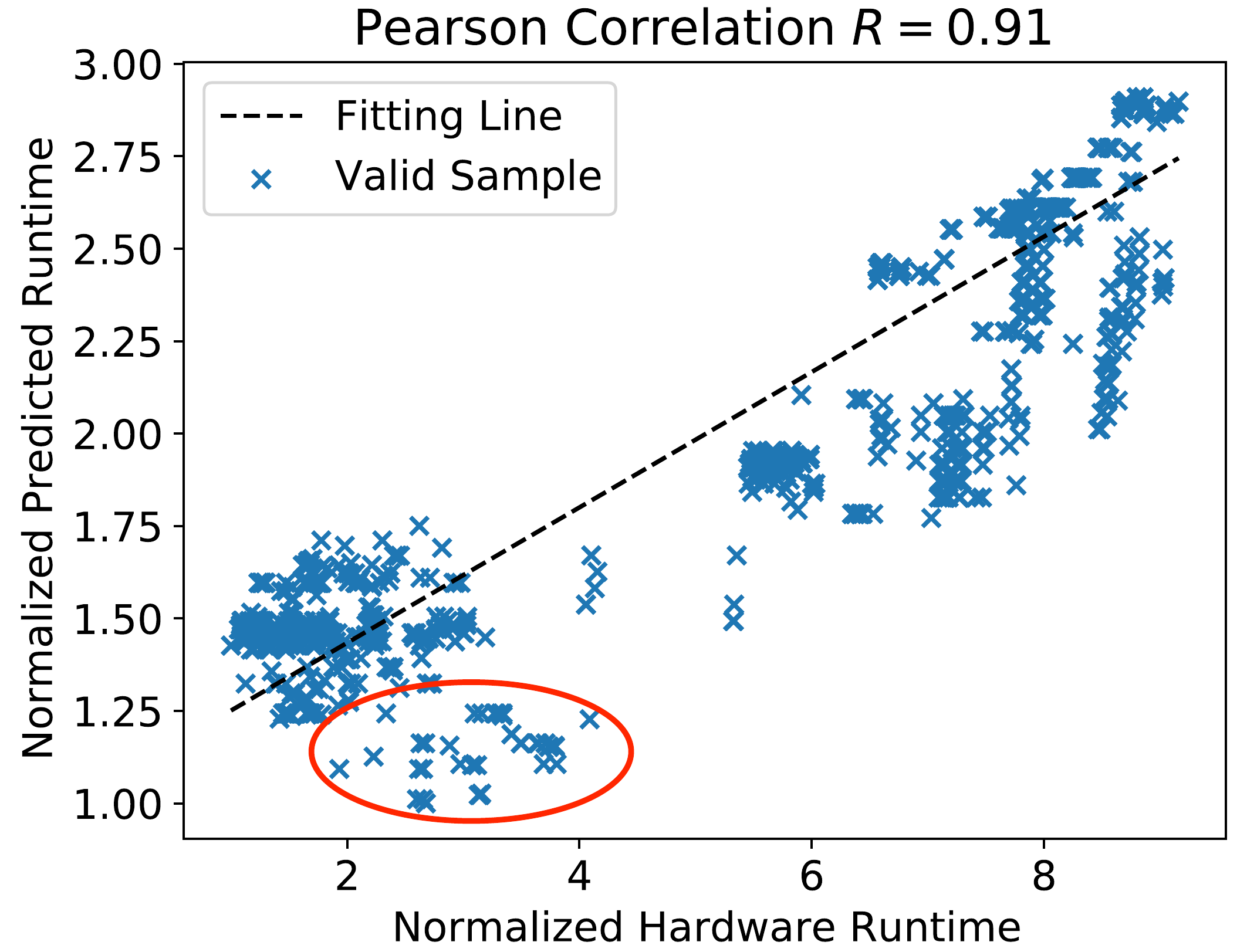}
    \caption{The hardware runtime and predicted runtime of valid partition samples for BERT model normalized to the minimal values of all samples respectively.}
    \label{fig:bert_regression}
\end{figure}

\subsection{Analytical Cost Model Accuracy} \label{sec:cost_model}
To justify the use of an analytical cost model during pretraining, we conduct a calibration study on the analytical model.
We randomly generate 2000 samples on BERT. Then we evaluate these 2000 samples on real hardware.
Finally, we normalize the predicted runtime and measured runtime to their minimal respectively.
The runtime is defined as the maximum latency across all chips, which is the reciprocal of the throughput. 
Figure~\ref{fig:bert_regression} shows the normalized predicted runtime vs. the normalized measured runtime for all valid samples from these 2000 samples.
We have three main observations from this calibration study:
1) 13.5\% of generated partitions are invalid on real hardware. 2) Some partitions showing lower runtime do not work well on real hardware, such as samples from the red cycle in Figure~\ref{fig:bert_regression}. 3) There is a strong correlation (Pearson correlation $R=0.91$) between predicted runtime and measured runtime.

Observing the critical features of the analytical cost model, we conclude that RL zero-shot does not work well for BERT on real chips because dynamic constraints cannot be captured by the constraint solver (13.5\% failures) and there are some false-positive results. 
However, RL fine-tuning can start from a pre-trained model efficiently due to the strong correlation between the results of the analytical cost model and that of the real hardware evaluation. 
The knowledge created during the pre-training with the majority of accurate samples can be still successfully transferred. 
For example, the knowledge of balanced placement and chip static constraints can be transferred.

\section{Conclusion}

In this work, we develop a deep RL solution working with a constraint solver for ML model partitioning targeting an MCM package.
Our method is generalizable to unseen input graphs via a pre-training pipeline.
Our evaluation of a production-scale model, BERT, on real hardware evaluation shows that our approach can outperform random search and simulated annealing by 6.11\% and 5.85\% at convergence.
Finally, our RL-based approach is transferable. The fine-tuning on a pre-trained model improves the sample efficiency up to 21.15x than RL training from scratch on BERT placement. This effectively reduces the search time from more than 3 hours to 9 minutes to achieve a throughput improvement of 2.6x, compared to a greedy heuristic in the production compiler.


\bibliography{neurips2021}
\bibliographystyle{mlsys2022}

%


\end{document}